\pdfoutput=1

\documentclass[11pt]{article}

\usepackage[final]{acl}

\usepackage{times}
\usepackage{latexsym}

\usepackage{amsmath}
\usepackage{pifont}
\newcommand{\cmark}{\ding{51}}%
\newcommand{\xmark}{\ding{55}}%

\usepackage{graphicx}
\usepackage{booktabs}
\usepackage[ruled,vlined]{algorithm2e}

\usepackage[T1]{fontenc}

\usepackage[utf8]{inputenc}

\usepackage{microtype}

\usepackage{inconsolata}

\usepackage{graphicx}

\newcommand{\shortname}{BESC}
\newcommand{\longname}{Beam Search-based Example Sequence Constructor}

\title{Learning to Search Effective Example Sequences for In-Context Learning}

\author{Xiang Gao, Ankita Sinha, Kamalika Das \\
  Intuit AI Research \\
  2700 Coast Avenue, Mountain View, CA 94043 \\
  \texttt\{xiang\_gao, ankita\_sinha2, kamalika\_das\}@intuit.com}

\begin{document}
\maketitle

\begin{abstract}

Large language models (LLMs) demonstrate impressive few-shot learning capabilities, but their performance varies widely based on the sequence of in-context examples. Key factors influencing this include the sequence's length, composition, and arrangement, as well as its relation to the specific query. Existing methods often tackle these factors in isolation, overlooking their interdependencies. Moreover, the extensive search space for selecting optimal sequences complicates the development of a holistic approach.
In this work, we introduce \longname~(\shortname), a novel method for learning to construct optimal example sequences. \shortname~addresses all key factors involved in sequence selection by considering them jointly during inference, while incrementally building the sequence. This design enables the use of beam search to significantly reduce the complexity of the search space. Experiments across various datasets and language models show notable improvements in performance.\footnote{The code will be released at \url{https://github.com/intuit-ai-research/BESC}.}

\end{abstract}
\section{Introduction}

\begin{figure}[htbp]
    \centering
    \includegraphics[width=7.5cm]{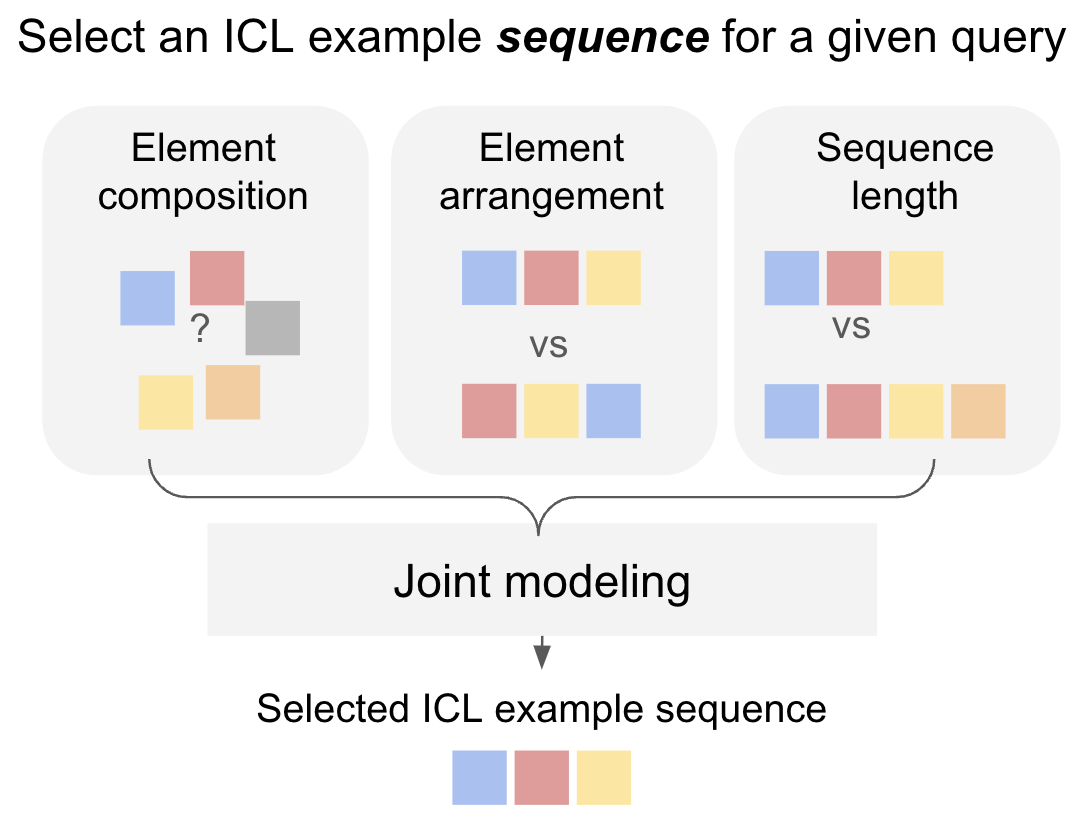}

\caption{
The in-context learning (ICL) performance of large language models (LLMs) depends on the length, composition, and arrangement of example sequences. Existing methods address these factors separately, while our algorithm jointly considers them and efficiently manages the search space with beam search.
}

    \label{fig_intro}
\end{figure}

Large language models (LLMs) have demonstrated impressive few-shot learning capabilities \citep{brown2020language}, where they can learn to provide better responses from just a few examples provided in the prompt. This in-context learning (ICL) ability  \cite{brown2020gpt3, gao2024customizing} has found a wide range of applications. 
However, LLMs may not always truly understand or generalize from the few-shot examples provided \citep{min2022rethinking, wei2023differently}, and their few-shot 
performance is highly sensitive to the \emph{sequence} of examples used \citep{yizhe2021fewshot, lu2021order}. Therefore, the selection of the example sequence becomes an important factor in leveraging LLMs' ICL capabilities. 
Since the fundamental mechanism behind in-context learning remains unclear \citep{min2022rethinking, xie2021explanation, olsson2022context}, existing work often approaches example sequence selection either by using heuristics or by focusing on specific subproblems. 
There is a lack of approaches that holistically learn to select the optimal example sequence, covering different aspects of the problem.

\begin{table*}
    \centering
    \small
    \begin{tabular}{cccccc}
        \toprule
        & Dynamic & Seq. Len. & Composition & Arrangement & \\
        \hline
       kNN \citep{yizhe2021fewshot} & \cmark & \xmark & \cmark & \xmark & \\
       EPR \citep{rubin2021epr} & \cmark & \xmark & \cmark & \xmark & \\
       PromptPG \citep{lu2022dynamic} & \cmark & \xmark & \cmark & \xmark & \\
       Cover-LS \citep{levy2022diverse} & \cmark & \xmark & \cmark & \xmark & \\   
       CEIL \citep{ye2023compositional} & \cmark & \xmark & \cmark & \xmark & \\    
       Reordering \citet{lu2021order}  & \xmark  & \xmark & \xmark & \cmark  & \\
       Q-learning \citep{zhang2022active} & \xmark  &  \cmark & \cmark & \cmark & \\
        \hline
       \shortname~(Ours)  &  \cmark  &  \cmark  &  \cmark &  \cmark  & \\
        \bottomrule
    \end{tabular}

    \caption{
Existing methods address different aspects of example sequence selection, such as query dependence (Dynamic selection), sequence length (Seq. Len), example composition (Composition), and example order in the prompt (Arrangement). Our method considers all these factors.
}

    \label{table-checkmarks}
\end{table*}
 
As illustrated in Figure~\ref{fig_intro}, the example sequence selection problem involves multiple factors, including the dependence on specific queries \citep{yizhe2021fewshot}, composition \citep{ye2023compositional, levy2022diverse}, arrangement \citep{lu2021order}, and the number of elements \citep{zhang2022active}. These factors should be jointly considered to achieve optimal performance. For example, \citet{zhang2022active} showed that without good element composition, merely reordering may not improve performance.

However, most existing methods focus on isolated subproblems without addressing the interdependence of different aspects of sequence selection, as illustrated in Table~\ref{table-checkmarks}. \citet{rubin2021epr} and \citet{lu2022dynamic} overlook the relationships between examples within the sequence, selecting examples independently. On the other hand, \citet{levy2022diverse} and \citet{ye2023compositional} consider the interactions between examples but neglect the effects of ordering and the number of examples. \citet{zhang2022active} build the sequence incrementally, but they rely only on the current sequence length as a feature to select the next example, leading to the loss of valuable textual information and limiting the method's applicability to dynamic example selection.

Most of the existing methods rely on researchers' intuition rather than on learning algorithms. This is likely due to the enormous search space of the example sequence selection problem. The number of possible sequences grows exponentially with the number of example candidate and potential sequence length. Consequently, existing learning-based example sequence algorithms either treat examples independently \citep{rubin2021epr,lu2022dynamic} or use simplified sequence representations \citep{zhang2022active}.

In this work, we propose a novel approach, \longname~(\shortname), to tackle the example sequence selection problem using a learning algorithm that jointly considers all aspects of the sequence—dynamically selecting examples for each query while accounting for composition, arrangement, and sequence length in a holistic manner. The model is designed to allow sequence construction incrementally at inference time, which makes it possible to employ a beam search algorithm to address the challenge of huge search space complexity. 

\begin{figure*}[htbp]
    \centering
    \includegraphics[width=13cm]{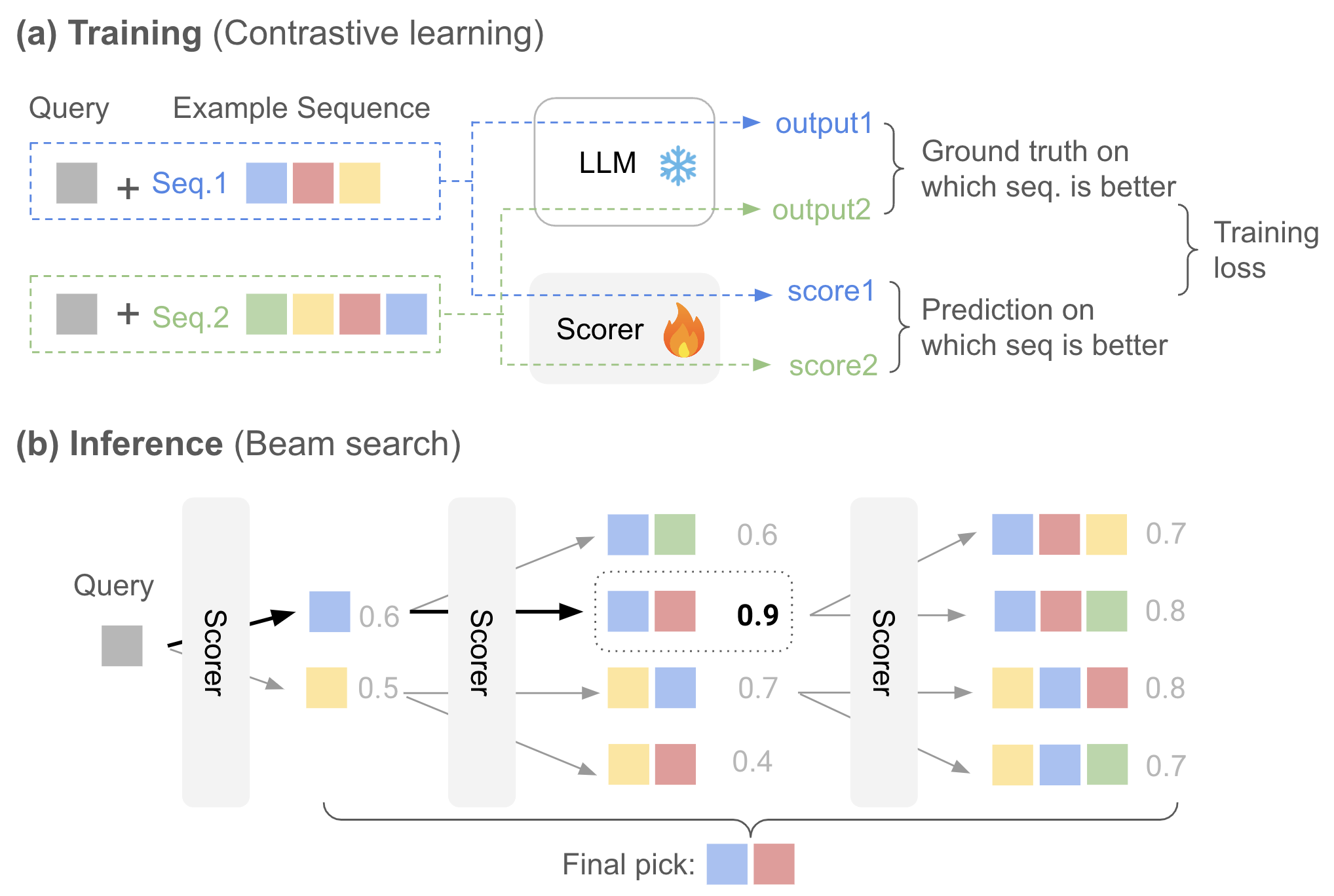}

    \caption{
An overview of our \shortname~approach. 
(a) We train a scoring model to predict the effectiveness of an example sequence for a given query, using contrastive learning to compare LLM performance between sequences.
(b) During inference, the example sequence is constructed incrementally with beam search, where the trained scorer ranks candidates and prunes nodes.
}

    \label{fig_method}
\end{figure*}
\section{Background}

\subsection{In-Context Learning Mechanisms}

In-context learning (ICL) refers to the ability of models to learn tasks by using only a few examples as demonstrations, without updating their parameters \cite{brown2020language}. Since ICL is training-free, it greatly reduces the computational cost of adapting models to new tasks. However, the mechanisms underlying ICL remain unclear.

While \citet{min2022rethinking} suggest that certain language models primarily rely on semantic prior knowledge triggered by the examples, \citet{wei2023differently} show that large models can override semantic priors when presented with in-context exemplars that contradict these priors. This suggests that both semantic priors and input–label mappings play important roles in ICL. Additionally, \citet{olsson2022context} argue that the ICL capabilities of large language models are rooted in "induction heads," which are transformer \cite{vaswani2017attention} circuits formed during pretraining that can copy patterns from ICL examples. On the other hand, \citet{xie2021explanation} showed that not only transformers but also LSTM models \cite{hochreiter1997lstm} can exhibit ICL behavior.

The unclear nature of ICL mechanisms, along with its dependence on model scale and architecture, underscores the potential for learning-based approaches to optimize the selection of example sequences accordingly.

\subsection{Example Sequence Selection}

Suppose we have a set of examples for a given task, ${e_i}$, as a candidate pool, where each example is a pair consisting of an input and a label, $e_i=(x_i, y_i)$. The example sequence selection problem involves selecting a sequence, $E=[e_1, e_2, \dots, e_k]$, to include in the prompt in order to improve the ICL generation quality. The selection may depend on the query $x^\text{query}$ (dynamic examples) or may be independent of it (static examples). Example sequence selection is typically studied across three main aspects: the composition of the examples, their arrangement, and the length of the sequence. A comparison of existing methods with respect to these factors is shown in Table~\ref{table-checkmarks}.

\paragraph{Static examples} 
In this setting, the goal is to select $E$ for a given task, and this selection does not depend on the query $x^\text{query}$. In practice, researchers use a set of labeled queries as a validation set to determine $E$ based on specific metrics evaluated on this set \cite{lu2021order, zhang2022active}.

\paragraph{Dynamic examples} 
In this setting, the goal is to dynamically select a sequence of ICL examples, $E$, for each given query $x^\text{query}$. The dynamical selection is expected to improve performance compared to using a static sequence, as different queries—even within the same task—may require different sets of skills and information demonstrated by varying examples. Several works have explored this approach \cite{yizhe2021fewshot, an2023skill, rubin2021epr, lu2022dynamic, levy2022diverse, ye2023compositional}.

\paragraph{Example composition}
This line of work focuses on the composition of examples. \citet{yizhe2021fewshot} proposed selecting the top $k$ examples, ${e_i}$, that are semantically similar to $x^\text{query}$. \citet{an2023skill} suggested selecting examples that demonstrate a similar set of "skills" required to solve $x^\text{query}$. Beyond similarity, diversity is another important factor in determining example composition. \citet{levy2022diverse} proposed selecting diverse examples that cover different aspects needed to solve $x^\text{query}$. \citet{ye2023compositional} jointly considered both similarity and diversity in example selection. Additionally, \citet{zhang2022active} and \citet{lu2022dynamic} employed reinforcement learning to optimize example selection.

\paragraph{Example arrangement}
\citet{lu2021order} demonstrated that the order in which examples are presented in the prompt significantly impacts LLM performance. They proposed using several entropy-based metrics, measured on a validation set, to determine the optimal order. These metrics are grounded in the intuition that the model's predictions should not be overly confident or excessively unbalanced.

\paragraph{Sequence length} 
LLMs may struggle to effectively utilize long contexts \cite{liu2024lost}, and ICL performance does not necessarily improve as more examples are provided. Therefore, sequence length is another important factor to consider in example sequence selection. \citet{zhang2022active} proposed including an "early termination" action when constructing the example sequence to control its length and prevent performance degradation.

\begin{figure*}[htbp]
    \centering
    \includegraphics[width=14cm]{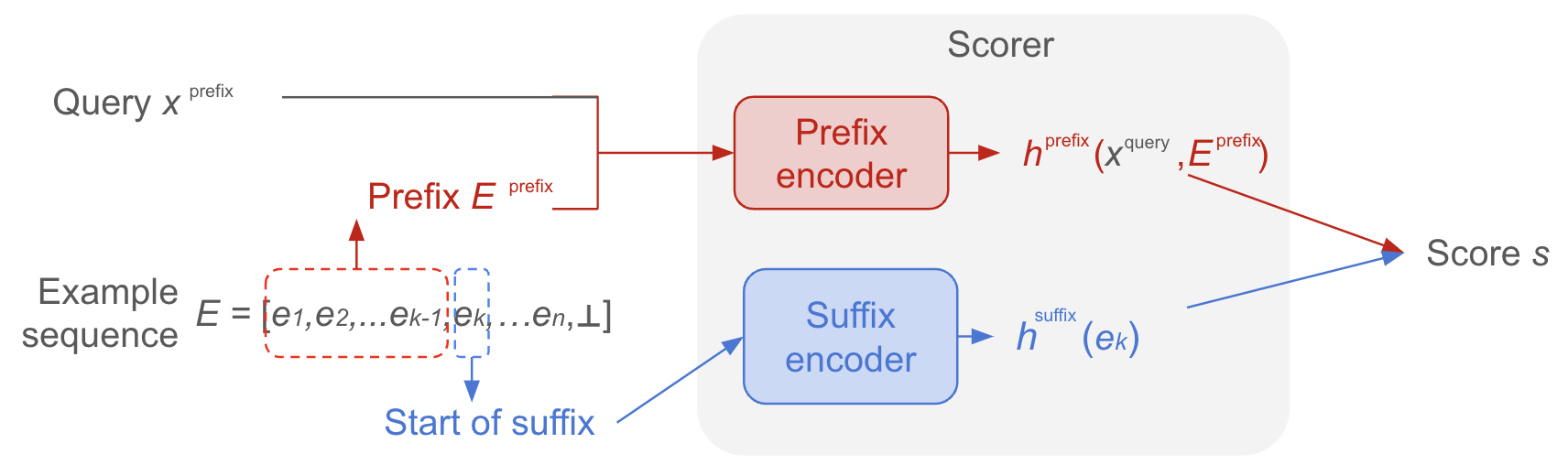}

    \caption{
An illustration of the dual-encoder architecture of the scoring model. This design enables the model to rank the next element for a given query and prefix, allowing incremental sequence construction with beam search during inference.
}

    \label{fig_model}
\end{figure*}

\section{Method}
\label{sec-method}

We propose \longname~(\shortname) to address the challenges of holistically multiple aspects of example sequence selection and the huge search space complexity.

To learn how to construct the optimal example sequence, we train a model to score example sequences for a given query. The scoring model $s(E, x^\text{query})$ is trained to predict the relative LLM generation quality given the query $x^\text{query}$ and example sequence $E$. Using this scoring model as a guiding signal, we employ a beam search algorithm to navigate the vast sequence space at inference time. Figure~\ref{fig_method} provides an illustration of the training and inference stages of our approach.

\subsection{Model Design}

We model the sequence using a prefix-suffix decomposition, enabling incremental construction during inference. At each step, the query and the current prefix $E^\text{prefix} = [e_1, e_2, \dots, e_{k-1}]$ are used to select the next element $e_k$, or the head of the suffix. This element is appended to the prefix, and the process repeats until a special termination element, $\perp$, is selected, marking the end of the sequence.

To enable this prefix-suffix selection, we design a dual-encoder architecture (Figure~\ref{fig_model}). The model generates an embedding for the prefix, $h^{\text{prefix}}(x^\text{query}, E^\text{prefix})$, and another for the suffix, $h^{\text{next}}(e_k)$. The score $s$ is calculated as the dot product of these two embeddings:
\[
s(E, x^\text{query}) = h^{\text{prefix}}(x^\text{query}, E^\text{prefix}) \cdot h^{\text{suffix}}(e_k)
\]

Since the remaining elements $[e_{k+1}, \dots, e_l]$ are unknown at inference time, the suffix encoder only considers the head of the suffix, $e_k$. This forces the model to learn to predict the effectiveness of a sequence without having access to all its elements. However, the model can also score a complete sequence by setting the suffix to the termination element $\perp$, with the prefix comprising the actual sequence elements $[e_1, e_2, \dots, e_l]$.

The dot product formulation transforms this ranking task into a nearest-neighbor search in the embedding space, which can be efficiently performed using a vector store with precomputed embeddings (see Section~\ref{sec-infer}).

\subsection{Model Training}

\begin{algorithm}[t]
\small
\newcommand\mycommfont[1]{\footnotesize\ttfamily\textcolor{gray!50}{#1}}
\SetCommentSty{mycommfont}

\textbf{Requirement:} \\
$\{e^{\text{cand}}_i\}$ the example candidate pool.

$\{e^{\text{query}}_i\}$ the set of labeled queries.

$L$ the maximum length of example sequence.

\textbf{Training:} 
    
\mycommfont{// Illustrated with batch size 1 in 1 epoch}

\For{$e^\text{query}_i = (x^\text{query}_i, y^\text{query}_i)$ in $\{e^{\text{query}}_i\}$}{

    \mycommfont{// Sample a prefix sequence}

    $l^\text{prefix} \gets$ Rand. int. $0 \leq l^\text{prefix} \leq L$

    $E^\text{prefix} \gets$ $l^\text{prefix}$ rand elem from $\{e^{\text{cand}}_i\}$
    
    \mycommfont{// Sample two suffix sequences}
        
    \For{$j \in \{1, 2\}$}{

        $l_j \gets $ Rand int $1 \leq l^\text{prefix} + l^\text{suffix}_j \leq L$
    
        $E^\text{suffix}_j \gets$ $l^\text{suffix}_j$rand elem from $\{e^{\text{cand}}_i\}$

        $E_j \gets E^\text{prefix} + E^\text{suffix}_j$
    
        $g_j \gets \text{LLM}(x^\text{query}_i, E_j)  $
    
        $a_j \gets$ Quality of $g_j$ given $y^\text{query}_i$
        
        $s_j \gets s_\theta(x^\text{query}_i, E_j )$
        
        }

        \mycommfont{// Which is a better example sequence for $x^\text{query}_i$? The ground truth:}
        
        $q \gets \begin{cases} [1,0] & \text{if } a_1 > a_2 \\ [0,1] & \text{otherwise} \end{cases}$

        \mycommfont{// and the prediction:}

        $p \gets \text{Softmax}([s_1, s_2])$

        \mycommfont{// Loss and optimization}

        $L \gets \text{CrossEntropyLoss}(p, q)$

        Update $\theta$ given $\nabla_\theta L$
}

\Return $s_\theta$
\caption{Training \shortname scorer}
\label{algo-train}
\end{algorithm}

We train the scoring model to rank example sequences using a contrastive learning approach, as outlined in Algorithm~\ref{algo-train}.

For a given task, the training dataset is divided into an example candidate set, $\{e_i^\text{cand}\}$, and a query set, $\{e_i^\text{query}\}$. The maximum number of examples in the sequence is $L$, a hyperparameter. We assume access to an automatic evaluation metric, $a(g,y)$, that measures the quality of the LLM's generated output $g$ against a reference label $y$\footnote{The metric can also be reference-free, provided it yields a scalar quality score.}.

For each query, we sample a pair of example sequences, $E_1$ and $E_2$, which share a prefix $E^\text{prefix}$, but their suffix may differ in length, composition, and/or arrangement. The prefix $E^\text{prefix}$ could be empty, in which case the next element selected will be the first in the sequence.

The goal is to train the model to predict which of the two sequences will lead the LLM to generate a higher-quality output for the given query $x_i^\text{query}$. The ground truth ranking is determined by evaluating the LLM's output for each sequence $E_j$ using the evaluation metric $a$.

We optimize the model by minimizing the cross-entropy loss between the ground truth ranking and the predicted probabilities. In this contrastive learning setup, the sequence with better output quality is treated as the positive sample, and the other as the negative sample.

\subsection{Inference with Beam Search}
\label{sec-infer}

\begin{algorithm}[t]
\small
\newcommand\mycommfont[1]{\footnotesize\ttfamily\textcolor{gray!50}{#1}}
\SetCommentSty{mycommfont}

\textbf{Requirement:} \\

$x^\text{query}$ the query input.

$\{e^{\text{cand}}_i\}$ the example candidate pool.

$L$ the maximum sequence length.

$s_\theta$ the trained scoring model

$b$ the beam width

$c$ the number of next elements to consider for each prefix

Trained model $s_\theta(E, x^\text{query})$ to rank candidates

\textbf{Beam search:} 

$B^0 \gets \{\emptyset\} $, initial  beam
    
\For{$l \gets 1 \text{ to } L$}{
    
    \mycommfont{// Prune previous beam with scorer}

    $B^{l-1}_\text{kept} \gets$ Top $b$ candidates in $B^{l-1}$ 

    \mycommfont{// Branch and save in the new beam}
    
    $B^l \gets \emptyset $, current beam

    \For{$E_i \in B^{l-1}_\text{kept}$}{
        
        ${e^\text{next}_j} \gets $ Top-$c$ candidates given $E_i$
        \For{$j \gets 1 \text{ to } c$}{
            
            Add seq $[E_i,e^\text{next}_j]$ to $B^l$
            
        }
    }
}

\Return $E$ of the highest \shortname~score $s_\theta(x,E)$
\caption{\shortname~inference}
\label{algo-infer}
\end{algorithm}

The search space for example sequences is vast, on the order of $O(N^L)$, where $N$ is the number of example candidates and $L$ is the maximum sequence length\footnote{The complexity to evaluate all possible sequences from length 1 to $L$ is $\sum_{k=1}^{L} A(N, k) = \sum_{k=1}^{L} \frac{N!}{(N-k)!}$, which can be approximated as $N^L$ when $N$ is much larger than $L$.}. An efficient search algorithm is therefore necessary.

Inspired by its success in speech understanding \cite{medress1977speech}, natural language generation \cite{och1999improved}, and recommendation systems \cite{gao2020deepretrieval}, we employ beam search to construct the example sequence during inference. Beam search is a form of pruned breadth-first search where the breadth is limited by a parameter, $b$, called the beam width.

We incrementally extend the example sequence by appending promising examples to the partial sequence while pruning less promising ones. The algorithm is detailed in Algorithm~\ref{algo-infer}.

At each step, we use the embedding $h^\text{prefix}(x^\text{query}, E^\text{prefix})$ to find the top $c$ examples with the nearest embeddings, $h^\text{suffix}(e_i)$, that have not yet been added to $E^\text{prefix}$. This step can be efficiently implemented using a vector store with pre-computed $h^\text{suffix}(e_i)$ embeddings, using the dot product as the distance measure. This process generates $b \cdot c$ example sequences at each time step, which are pruned by the scoring model, retaining only the top $b$ candidates as the new prefix for the next step.

Initially, $E^\text{prefix}$ is empty, so the first step involves selecting the most promising starting example for the given query. We repeat the grow-prune cycle until a sequence is pruned, reaches the maximum length $L$, or the end action $\perp$ is selected.

Finally, we rank all complete sequences explored during beam search, which may vary in length, based on their score $s$, returning the top-ranked sequence as the final selection (see Figure~\ref{fig_method}).

\subsection{Reduced Complexity}

Pruning allows us to score only a small subset of all possible example sequences, which greatly reduces the size of the search space.

In addition, our dual-encoder architecture, combined with the dot product formulation and approximate k-nearest neighbor (k-NN) search \cite{andoni2008near}, makes the process more efficient. The prefix encoder $h^\text{prefix}$ is run only once per prefix, and the suffix encoder $h^\text{suffix}$ is computed just once per candidate example, significantly lowering computational costs.

The overall complexity to score sequences is reduced from $O(N^L)$ to:
\[
O(L \cdot b \cdot c) + O(N) + O(L \cdot b \cdot c \cdot \log N)
\]
The first term corresponds to the computation of the prefix encoder for all prefix candidates explored. The second term accounts for computing the suffix encoder for all $N$ suffix head candidates. The third term represents the complexity of the approximate k-NN search to find the top $c$ examples at each step.

\begin{table*}[!ht]
\centering
\small
\begin{tabular}{c|ccc|cc|cc|cc}
\toprule

  & \multicolumn{3}{|c|}{GPT-2} &  \multicolumn{2}{|c|}{GPT-3} &  \multicolumn{2}{|c}{GPT-neo} &  \multicolumn{2}{|c}{GPT-3.5} \\
                                          & AGNews & SST2 & TREC & TabMWP  & BREAK & SST5  & BREAK & TREC & GSM8K \\
\hline
Random                                    & 0.55   & 0.66 & 0.41 & 0.65    & 0.04  & 0.31  & 0.02  & 0.55 & 0.87 \\
Reordering\tiny{ \citep{lu2021order} }    & 0.63   & 0.68 & 0.33 & -       & -     & -     & -     & 0.59 & 0.91 \\
kNN\tiny{ \citep{yizhe2021fewshot} }      & -      & -    & -    & 0.68    & -     & -     & -     & 0.70 & 0.90 \\
EPR\tiny{ \citep{rubin2021epr}}           & -      & -    & -    & -       & 0.25  & 0.43  & 0.30  & 0.75 & 0.93 \\
Q-learning\tiny{ \citep{zhang2022active}} & 0.71   & 0.81 & 0.43 & -       & -     & -     & -     & -    & -  \\
PromptPG\tiny{ \citep{lu2022dynamic}}     & -      & -    & -    & 0.71    & -     & -     & -     & -    & -  \\
CEIL\tiny{ \citep{ye2023compositional}}   & -      & -    & -    & -       & -     & 0.47  & 0.34  & -    & -  \\

\hline

\shortname\tiny{(ours)}                        & \textbf{0.77} & \textbf{0.87}  & \textbf{0.48} & \textbf{0.77} & \textbf{0.28} & \textbf{0.50} & \textbf{0.38} & \textbf{0.85} & \textbf{0.97} \\

\bottomrule
\end{tabular}
\caption{\label{table-strong-baseline}
Comparison between the accuracy achieved by our method (\shortname) and baseline methods.
}
\end{table*}
\section{Experiments}

\subsection{Datasets}
\label{sec-data}

Given the wide application of few-shot learning with LLMs, prior works on example sequence selection have been evaluated across a range of tasks, including text classification \cite{Zhang2015agnews, socher2013sst2, li-roth-2002-learning, hovy-etal-2001-toward} and text generation \cite{Wolfson2020Break, li2020mtop, lu2022dynamic}. 

Following this, we consider a broad set of six datasets\footnote{GSM8K and BREAK are released under the MIT license, and TabMWP under the CC BY-NC-SA license.}, including AGNews \cite{Zhang2015agnews}, SST \cite{socher2013sst2}, and TREC \cite{li-roth-2002-learning, hovy-etal-2001-toward} for text classification, and TabMWP \cite{lu2022dynamic}, BREAK \cite{Wolfson2020Break}, and GSM8K \cite{cobbe2021training} for text generation, as summarized in Table~\ref{table-data}.

\begin{table}
    \centering
    \small
    \begin{tabular}{ccc}
        \toprule
       Dataset & Task & Train set \\
        \hline
       AGNews  & Topic classification & 44.3k \\
       SST  & Sentiment classification & 67.3k \\
       TREC  & Question classification & 5.5k \\
       TabMWP  & Mathmatical reasoning & 38.4k \\
       BREAK & Meaning representations & 44.3k \\
       GSM8K & Mathmatical reasoning & 7.5k \\
        \bottomrule
    \end{tabular}
\caption{
Datasets used in this work.
}

    \label{table-data}
\end{table}

\subsection{Language Models}

We experiment with four language models of varying sizes and training paradigms: GPT-2 \citep{radford2019gpt2}, GPT-neo \citep{black2021gptneo}, GPT-3 \cite{brown2020gpt3}, and GPT-3.5.

While prior studies have primarily focused on earlier models that lack instruction tuning \cite{ouyang2022instructgpt}, we include GPT-3.5 to examine the effect of example sequence selection on instruction-tuned models.

Our experiments cover both small and large models, spanning open-source models (GPT-2 and GPT-neo) and closed API models (GPT-3, GPT-3.5), as well as models that are pretrain-only and those with post-training (instruction-tuned) capabilities.

\subsection{Baselines}

We compare our method with several existing example selection approaches, including kNN \cite{yizhe2021fewshot}, EPR \cite{rubin2021epr}, CEIL \cite{ye2023compositional}, and PromptPG \cite{lu2022dynamic}, which focus on example composition. We also include Reordering \cite{lu2021order}, which prioritizes example arrangement, and a Q-learning-based method \cite{zhang2022active}, which considers sequence length, composition, and arrangement but overlooks query dependence and textual information. A comparison of the design features of these methods and ours is summarized in Table~\ref{table-checkmarks}.

\subsection{Implementation}

\subsubsection{Modeling}

The two encoders in the model can be implemented in various ways. In our implementation, we ensure that the prediction depends on the textual information from the query and examples, while considering sequence length, composition, and arrangement.

For the prefix encoder, we use a hierarchical approach. We first encode $x^\text{query}$ and each $e_i$ using transformers initialized with a pretrained sentence BERT model \cite{reimers2019sbert}, then pass the resulting embeddings to a two-layer LSTM network to generate the final sequence representation. This hierarchical structure helps handle long sequences, especially when $L$ is large.

The suffix encoder also uses transformer-based textual encoders initialized with sentence BERT. Both the transformer and LSTM parameters are fine-tuned during contrastive learning.

\subsubsection{Training}

For each task, we split the training data into 1,000 instances as the query set, with the remainder serving as the example candidate set. We use greedy sampling for LLMs generation.

To generate positive and negative samples for contrastive learning, we use $L=7$ and apply uniform random sampling to generate the variables (e.g., $l^\text{prefix}$, $l^\text{suffix}_j$, $E^\text{prefix}$, and $E^\text{suffix}_j$). 
However, more effective sequences for contrastive learning may be obtained by either leveraging existing example selection methods or using the \shortname~scoring model trained in the previous epoch. Adding these samples to the training data may further improve the model. We leave the exploration of such strategies for future work.

\subsection{Inference}

We perform beam search using $b=5$ beams and a maximum of $c=5$ candidates per beam. 

\section{Results and Discussion}

In this section, we present our results, comparing our method to baselines and analyzing key components through ablation studies. We also examine the model's transferability across tasks.

\subsection{Improved Performance over Baselines}

We compare our method with the baseline approaches on the datasets they were originally tested on. The results, summarized in Table~\ref{table-strong-baseline}, show that our method consistently outperforms all baselines across six datasets and four language models.

We hypothesize that this improvement is due to the holistic consideration of all key factors involved in dynamic example sequence selection that our method incorporates.

To further investigate, we conduct ablation studies to evaluate the contribution of each key factor. All ablation studies are conducted on the TREC and GSM8K datasets using GPT-3.5 as the LLM.

\subsection{Effects of Dynamic Examples}

The first key factor we explore is the impact of dynamic example selection, where the example sequence is adapted to each query rather than being fixed for all queries (static examples). Previous works by \citet{lu2021order} and \citet{zhang2022active} focused on static examples, and we hypothesize that adapting the example sequence to the query can improve performance.

To evaluate this, we create a ``static'' version of \shortname~by removing $x^\text{query}$ from the inputs of the prefix encoder, making the prediction (or next element selection) not dependent on the query. We train this static version of the scoring model and compare it to the original, query-dependent "dynamic" \shortname.

\begin{table}[!ht]
\centering
\small
\begin{tabular}{ccc}
\toprule

 & Dynamic & Static \\
\hline
TREC & 0.85 & 0.79 \\
GSM8K & 0.97 & 0.92 \\

\bottomrule
\end{tabular}
\caption{\label{table-ablation-dynamic}
Ablation study on the effects of dynamic example sequence selection using two \shortname~variants. The ``Static'' version is trained without taking queries as input to the scoring model.
}

\end{table}

The results, summarized in Table~\ref{table-ablation-dynamic}, show that the dynamic version significantly outperforms the static version, highlighting the importance of selecting examples dynamically based on the specific query.

\subsection{Effects of Automatic Sequence Length}

Our method automatically determines the optimal sequence length (``Auto $l$'') during inference.\footnote{This is achieved by: 1) learning to selecting the termination token $\perp$, and 2) picking the best sequence from explored candidates of varying lengths during beam search. See Section~\ref{sec-method}} To study the impact of this feature, we compare \shortname~with fixed-length versions ($l=1$, $l=3$, $l=5$), where only candidates of the specified length are considered during inference.

\begin{table}[!ht]
\centering
\small
\begin{tabular}{ccccc}
\toprule

 & Auto $l$ & $l=3$ & $l=5$ & $l=7$ \\
\hline
TREC & 0.85 & 0.80  & 0.83 & 0.84 \\
GSM8K  & 0.97 & 0.91  & 0.95 & 0.95 \\

\bottomrule
\end{tabular}

\caption{\label{table-ablation-len}
Ablation study on the effects of example sequence length. \shortname, which automatically determines the sequence length ($1 \leq l \leq 7$), is compared with variants using fixed sequence lengths.
}

\end{table}

The results, summarized in Table~\ref{table-ablation-len}, show that automatically determining sequence length leads to further performance improvements compared to simply increasing the sequence length. We hypothesize that this is because the LLM may struggle to fully utilize longer contexts, especially when they include less useful or distracting information \cite{liu2024lost}.

\subsection{Effects of the Element Arrangement}

The impact of element ordering has been discussed by \citet{lu2021order}, but their work focused on the ordering of randomly selected elements. In contrast, we study the effects of ordering for carefully selected examples as part of our ablation study.

\begin{table}[!ht]
\centering
\small
\begin{tabular}{ccc}
\toprule

 & \shortname & Shuffled \\
\hline
TREC & 0.85 & 0.84 \\\
GSM8K & 0.97 & 0.95 \\

\bottomrule
\end{tabular}
\caption{\label{table-ablation-ordering}
Ablation study on the effects of example arrangement in the ICL example sequence. ``Shuffled'' uses the same examples as \shortname~for each query, but in a shuffled order.
}

\end{table}

We randomly shuffle the example sequence generated by \shortname~for each query and present the experimental results in Table~\ref{table-ablation-ordering}. The results show a decrease in performance after shuffling, indicating that element arrangement remains a relevant factor in dynamic example sequence selection.

\subsection{Effects of Sequential Modeling}

\citet{zhang2022active} observed performance improvement from example sequence selection in smaller models like GPT-2, but noted that this improvement diminishes with larger models such as GPT-3. They hypothesized that this reduction is due to the emerging capabilities of larger LLMs. However, other works, including ours, experimenting with GPT-3 \cite{lu2022dynamic, rubin2021epr}, show that example sequence selection can still provide significant performance improvements in larger models.

We suspect that the diminished improvement observed by \citet{zhang2022active} is due to their method using an oversimplified representation of the sequence, relying only on sequence length as the feature. 

To test this, we implemented a variant of \shortname~where the prefix encoder only uses the prefix length as the feature, ignoring the textual information. The performance of this variant decreases significantly compared to the original \shortname, highlighting the importance of more comprehensive sequential modeling.

\subsection{Transferability}

Similar to other learning-based methods \cite{lu2022dynamic, zhang2022active, rubin2021epr}, our approach requires task-specific training, which introduces a one-time additional cost compared to learning-free methods such as kNN \cite{yizhe2021fewshot}. To mitigate this, we explore transfer learning by applying a pretrained \shortname~model to new tasks without additional task-specific training.

\begin{table}[!ht]
\centering
\small
\begin{tabular}{c|c|ccc}
\toprule
 &     & \multicolumn{2}{|c}{BESC} \\ 
 & kNN & Single-task & Pretrained \\
\hline
TREC & 0.70 & 0.85 & 0.82  \\
GSM8K & 0.90 & 0.97 & 0.94  \\

\bottomrule
\end{tabular}

\caption{\label{table-transfer}
Comparison of a learning-free method (kNN), the original \shortname, and a pretrained version of \shortname.
}

\end{table}

In a leave-one-out setting, we use five of the six datasets listed in Table~\ref{table-data} to train a ``pretrained'' \shortname~scoring model and test its performance on the remaining target task. The experimental results, with TREC and GSM8K as target tasks and GPT-3.5 as the LLM, are summarized in Table~\ref{table-transfer}. The results show that while the pretrained \shortname~model performs worse than one trained specifically on the target task, it still outperforms the learning-free kNN method \cite{yizhe2021fewshot}. This suggests that knowledge about optimal example sequence selection can transfer across tasks.

\section{Conclusion}

We introduce \longname~(\shortname) to address the example sequence selection problem. \shortname~incrementally constructs sequences during inference, using beam search to reduce search space complexity. Experiments across multiple datasets and language models show substantial performance improvements. Ablation studies highlight key contributions: dynamic example selection, automatic sequence length, element arrangement, and sequential modeling. We also explore \shortname's potential in transfer learning. Lastly, we discuss limitations and propose future directions, including open-ended tasks and integration with prompting strategies.
\section{Limitations}
The datasets used in this work mainly consist of short text and focus on tasks with a unique ground truth. It is unclear how this method, or in-context learning (ICL) in general, performs on more open-ended tasks like dialogue \cite{zhang2020dialogpt}, role-play \cite{sadeq2024mitigating} or creative writing, which warrants further exploration.

Moreover, the effectiveness of ICL examples can
vary with different prompting strategies, espacially those eliciting LLM reasoning abilities \cite{wei2022cot, yao2024tot, lightman2023let}. The interaction between example selection and such techniques remains an open question.

The proposed method could introduce bias in the generated content through the selection of examples, and like other large language model techniques, it could be misused for harmful content generation.

\bibliography{custom}

\end{document}